\documentclass[10pt, a4paper]{article}
\usepackage{lrec}
\usepackage{graphicx}
\usepackage{tabularx}
\usepackage{soul}

\usepackage{amssymb}
\usepackage{amsmath}
\usepackage{stmaryrd}
\usepackage{proof}

\usepackage{subcaption}
\usepackage{verbatim}

\usepackage{pgf}

\usepackage{tikz}
\usetikzlibrary{matrix}
\usetikzlibrary{calc}
\usetikzlibrary{arrows}

\usepackage{algpseudocode}
\usepackage{algorithmicx}

\usepackage{epstopdf}
\usepackage[latin1]{inputenc}

\usepackage{hyperref}
\usepackage{xstring}

\usepackage{tabularx}
\usepackage{booktabs}
\newcolumntype{m}{>{\hsize=.65\hsize}X}
\newcolumntype{s}{>{\hsize=.3\hsize}c}
\newcolumntype{u}{>{\hsize=.1\hsize}c}
\usepackage{multicol}
\usepackage{multirow}

\usepackage{floatrow}

\newcommand{\secref}[1]{\StrSubstitute{\getrefnumber{#1}}{.}{}}
\newcommand{\duck}{\includegraphics[scale=0.022]{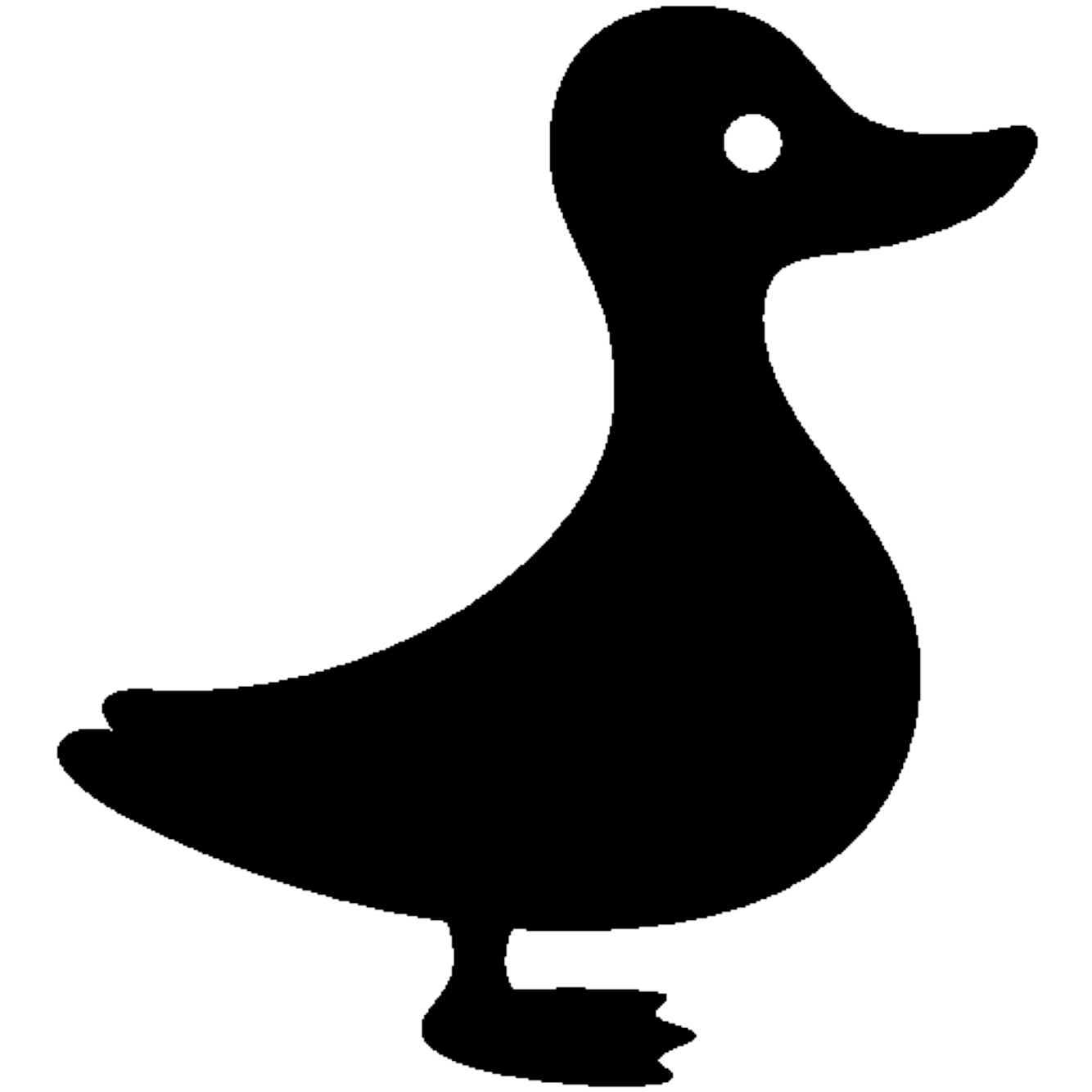}}
\newcommand{\flamingo}{\includegraphics[scale=0.020]{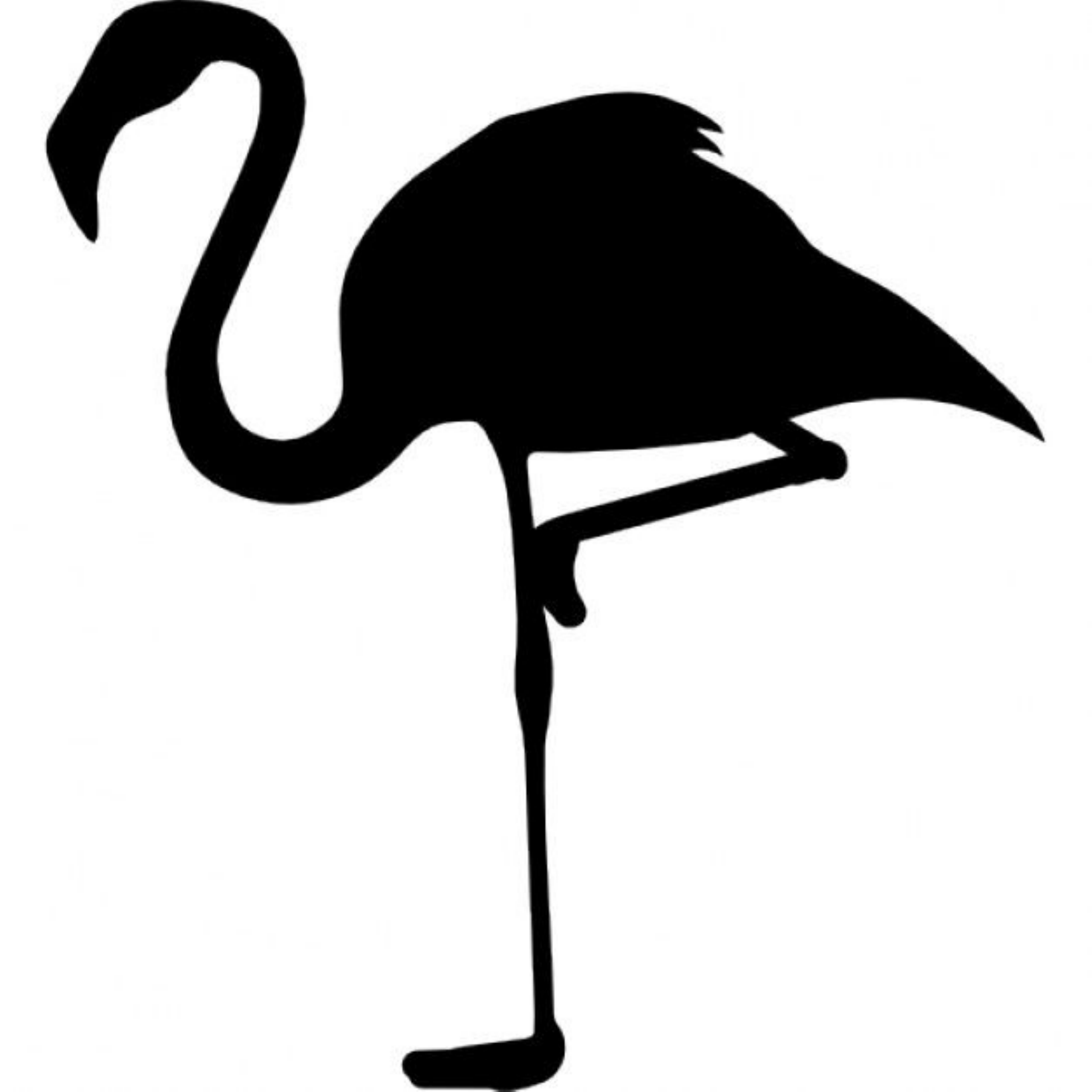}}
\newcommand{\li}{\multimap}
\newcommand{\type}[1]{\textsc{#1}}
\newcommand{\lolli}{\multimap}
\newcommand{\bang}{\mathop{!}}
\renewcommand{\Box}{\oblong}

\title{\AE THEL: Automatically Extracted Typelogical Derivations for Dutch}

\name{Konstantinos Kogkalidis\textsuperscript{\duck}, Michael Moortgat\textsuperscript{\duck}, Richard Moot\textsuperscript{{\flamingo}}}

\address{\textsuperscript{\duck}Utrecht Institute of Linguistics OTS, Utrecht University, \textsuperscript{\flamingo}LIRMM, Universit\'{e} de Montpellier, CNRS \\
        \{k.kogkalidis, m.j.moortgat\}@uu.nl, richard.moot@lirmm.fr \\
}

\abstract{
We present \AE THEL, a semantic compositionality dataset for written Dutch. 
\AE THEL consists of two parts.
First, it contains a lexicon of supertags for about 900\,000 words in context.
The supertags correspond to types of the simply typed linear lambda-calculus, enhanced with dependency decorations that capture grammatical roles supplementary to function-argument structures.
On the basis of these types, \AE THEL further provides 72\,192 validated derivations, presented in four equivalent formats: natural-deduction and sequent-style proofs, linear logic proofnets and the associated programs (lambda terms) for meaning composition.
\AE THEL's types and derivations are obtained by means of an extraction algorithm applied to the syntactic analyses of Lassy Small, the gold standard corpus of written Dutch.
We discuss the extraction algorithm and show how `virtual elements' in the original Lassy annotation of unbounded dependencies and coordination phenomena give rise to higher-order types.
We suggest some example usecases highlighting the benefits of a type-driven approach at the syntax semantics interface. 
The following resources are open-sourced with \AE THEL: the lexical mappings between words and types, a subset of the dataset consisting of 7\,924 semantic parses, and the Python code that implements the extraction algorithm.
}

\begin{document}

\maketitleabstract

\section{Introduction}
Typelogical categorial grammars offer an attractive view on the
syntax-semantics interface. The key idea is that the derivation
establishing the well-formedness of a phrase systematically
corresponds to a procedure that computes the meaning of that phrase,
given the meanings of its constituent words.

This connection between logical derivations and programs, known as the
Curry-Howard correspondence \cite{sorensen2006lectures}, plays a central role in
current typelogical grammars. In extended versions of the Lambek calculus \cite{lambek1958mathematics}
such as \cite{morrillTLG,DBLP:journals/jolli/Moortgat96,DBLP:journals/jolli/MorrillVF11,KubotaLevineMIT},
the Curry-Howard correspondence applies to a language of \emph{semantic} types, where meaning
composition is addressed. The semantic type language is obtained by a structure-preserving
translation from the \emph{syntactic} type system that handles surface word order, prosodic
structure and the like. Alternatively, the type system of Abstract Categorial Grammar~\cite{de2001towards}
and Lambda Grammars~\cite{Muskens2001-MUSGAT} is designed to capture a `tectogrammatical' \cite{curry1961some}
view of language structure that abstracts away from the surface realization. From the abstract syntax now
both the meaning representations and the surface forms are obtained by compositional translations, where for
the latter one relies on the standard encoding of strings, trees, etc in lambda calculus
\cite{barendregttypes}.

A common feature of these approaches is the use of \emph{resource-sensitive} type logics,
with Linear Logic~\cite{girard1987linear} being the key representative. Curry's original
formulation of the logic-computation correspondence was for Intuitionistic Logic,
the logic of choice for the formalization of mathematical reasoning. For many real-world
application areas, Linear Logic offers a more attractive alternative. In Linear
Logic, formulas by default stand for non-renewable resources that are used up in the course
of a derivation; free copying or deletion is not available.

The resource-sensitive view of logic and computation fits the nature of grammatical composition particularly well.
Despite this, linear logic is underrepresented when it comes to linguistic resources that exhibit its potential
at the syntax-semantics interface. Motivated by this lack of resources, we introduce
\AE THEL\footnote{\AE THEL: Automatically Extracted Theorems from Lassy}, a dataset of typelogical derivations for Dutch.
The type logic \AE THEL employs rests on the non-directional types of simply typed intuitionistic linear logic,
enhanced with modal operators that allow the inclusion of dependency information on top of simple function-argument relations.
\AE THEL consists of 72\,192 sentences, every word of which is assigned a linear type that captures its functional role within the phrase-local structure.
Under the parsing-as-deduction view, these types give rise to proofs, \textit{theorems} of the logic 
showing how the sentence type is derived from the types of the words it consists of.
These derivations are presented in four formats which, albeit being morally equivalent, can be varyingly suitable for different tasks,
viz.~proofnets, natural deduction and sequent-style proofs, and the corresponding
$\lambda$-terms.
\AE THEL is generated by an automated extraction algorithm applied to Lassy Small~\cite{vanNoord2013}, the gold standard corpus of written Dutch.
The validity of the extracted derivations is verified by LinearOne~\cite{Moot2017}, a mechanized theorem prover for first-order linear logic.

The structure of the paper is as follows: we begin in Section~\secref{section:type_logic}~by providing an overview of the type logic used. 
We proceed by detailing the proof extraction algorithm in Section~\secref{section:extraction}, before describing the dataset and its accompanying resources in Section~\secref{section:resources}. We give some conclusive remarks in Section~\secref{section:conclusion}.

\paragraph*{Related work} The work reported on here shares many of the goals of the CCGBank project~\cite{ccgbank}.

A difference is that CCGBank uses syntactic types aiming at wide-coverage parsing, whereas our linear types are geared
towards meaning composition. Also different is the treatment of long-range dependencies, where CCG establishes the 
connection with the `gap' by means of type-raising and composition rules. From a logical point of view, these
CCG derivations are suboptimal: they contain detours that disappear under proof normalization. In our
approach, long-range gaps are hypotheses that are withdrawn in a single inference step of conditional 
reasoning.

\section{Type Logic}
\label{section:type_logic}

\subsection{ILL${}_{\lolli}$}
The core of the type system used in this paper is ILL$_\lolli$, the simply-typed fragment of
Intuitionistic Linear Logic.
ILL$_\lolli$ types are defined by the following inductive scheme:

\[\type{T} ::= \type{A} \ | \ \type{T}_1 \li \type{T}_2\]

where $\type{A} \in \mathcal{A}$ is a set of \textit{atomic} types and linear implication ($\lolli$) is the
only type-forming operation for building complex types.
Atomic types are assigned to phrases that can be considered complete in themselves.
A complex type $\type{T}_1 \li \type{T}_2$ denotes a function that takes an argument of type $\type{T}_1$
to yield a value of type $\type{T}_2$. Such function types are \emph{linear} in the sense
that they actually consume their $\type{T}_1$ argument in the process of producing a
$\type{T}_2$ result.

Given Curry's ``proofs as programs'' approach, one can view the inference rules that determine
whether a conclusion $\type{B}$ can be derived from a multiset of assumptions $\type{A}_1,\ldots,\type{A}_n$ as the
\emph{typing rules} of a functional programming language. The program syntax is given by the
following grammar:

\[ M,N ::=  x \mid M N \mid \lambda x.M \]

Typing judgements take the form of sequents $$x_1:\type{A}_1,\ldots,x_n:\type{A}_n\vdash M:\type{B}$$ stating that
a program $M$ of type $\type{B}$ can be constructed with parameters $x_i$ of type $\type{A}_i$. Such a program,
in the linguistic application envisaged here, is a procedure to compute the meaning of a
phrase of type $\type{B}$, given meanings for the constituting words $w_i$ as bindings for
the parameters $x_i$.
The typing rules then comprise the Axiom $x:\type{A}\vdash x:\type{A}$, together with inference rules for the
elimination and introduction of the linear implication, where ($\lolli E$) corresponds
to function application, and ($\lolli I$) to function abstraction.

\begin{equation}
\label{eqn:elimination}
\infer[\li  E]{\Gamma,\Delta\vdash (M\ N): \type{B}}{\Gamma\vdash M: \type{A}\li \type{B} & \Delta\vdash N: \type{A}}
\end{equation}

\begin{equation}
\label{eqn:introduction}
    \infer[\li I]{\Gamma \vdash \lambda x.M: \type{A}\li \type{B}}{\Gamma, x: \type{A} \vdash M: \type{B}}\\
\end{equation}

The abstraction rule ($\lolli I$) comes into play in reasoning about higher-order types. 
We say that atomic types are of order zero. For function types, $\mathit{order}(\type{T}_1 \li \type{T}_2)$
is defined as $\textrm{max}(\mathit{order}(\type{T}_1)+1,\mathit{order}(\type{T}_2))$. A type of order $n>1$
is a function that has a function as one of its arguments. A relative pronoun, for example, could be
typed schematically as $(\type{NP}\li\type{S})\li\type{Rel}$. The import of this type is that the
relative clause body is an $\type{S}$ composed with the aid of an $\type{NP}$ hypothesis (the `gap');
this hypothesis is withdrawn by the ($\lolli I$) rule.

\subsection{Dependency enhancement}
As it stands, our type system doesn't have the expressivity to
capture \emph{grammatical role} information. For example, a ditransitive verb such as `give' would be typed as
$\type{NP}\lolli \type{NP}\lolli \type{NP}\lolli \type{S}$\footnote{Type brackets can be ommitted reading the $\lolli$ operation as right-associative.}; we would like to
be able to distinguish the subject, direct object and indirect object among the three $\type{NP}$ arguments.

To address this limitation, we use the modalities of Multimodal Categorial Grammars \cite{DBLP:journals/jolli/Moortgat96}.
Modalities are unary type-forming operations, commonly used for syntactic control, licensing
restricted forms of restructuring or reordering, or blocking overgenerating applications of such operations.
Here instead, we use them as a means of injecting dependency information directly into the type system
in the form of feature decorations\footnote{For earlier applications of modalities-as-features,
see \cite{heylen99,Johnson99}.}.

We augment the type system with unary operators $\diamond^d$ and $\Box^d$, where the $d$ labels are taken
from the set of dependency roles (subject, object, modifier, etc) in the corpus. 
The operators $\diamond^d$ and $\Box^d$ come with their own Elimination and Introduction rules. 
We assume that the dependency annotations do not affect the Curry-Howard program terms associated with a proof, so we can formulate these inference rules purely on the type level. 
For our purposes, $\diamond^d$ Introduction and $\Box^d$ Elimination play a key role.

\begin{equation}
\label{eqn:dia_intro_box_elim}
    \infer[\diamond I]{\langle \Gamma \rangle^d \vdash \diamond^d \type{A}}{\Gamma \vdash \type{A}} \qquad
    \infer[\Box E]{\langle \Gamma \rangle^d \vdash \type{A}}{\Gamma \vdash \Box^d\type{A}}
\end{equation}

The ($\diamond^d I$) rule says that if from resources $\Gamma$ one can derive a phrase of
syntactic type $\type{A}$, then one can obtain an $\type{A}$ phrase with dependency role $d$ by grouping
together the $\Gamma$ resources with the delimiter $\langle\cdot\rangle^d$, indicating
that they consitute a \emph{dependency domain}. The ($\Box^d E$) rule, similarly, 
encloses the $\Gamma$ resources with the dependency delimiter $\langle\cdot\rangle^d$,
this time by unpacking the $\Box^d A$ type of the premise, so that reasoning can now
proceed with the $A$ subtype.

The  $\diamond^d$ and $\Box^d$ modalities allow us to reconcile the demands of
the logical function-argument structure of a phrase with those of its dependency structure~\cite{depstruct}.

(i) We use $\diamond^d\type{A}\li\type{B}$ for a function acting as a \emph{head} that selects
for an $\type{A}$ \emph{complement} and assigns it the dependency role $d$ by means of ($\diamond^d I$).
For example, intransitive verbs are typed $\diamond^{\text{su}}\type{NP}\li\type{S}$.

(ii) We use $\Box^d(\type{A}\li\type{B})$ for \emph{non-head} functions, i.e.~functions
that, in the dependency structure, are dependents with respect to their argument.
For example, $\Box^{\text{mod}}(\type{NP}\li\type{NP})$ for a noun-phrase modifier, 
or $\Box^{\text{det}}(\type{N}\li\type{NP})$ for a determiner.
 
To complete the tour of the type system, the \emph{nested} implications of higher-order types
come with a structural modality $\bang$ on their argument: $\bang\type{A}\li\type{B}$.
The purpose of the $\bang$ modality is to allow for the $\type{A}$ hypothesis that is
withdrawn in the ($\li I$) step to originate from an embedded dependency domain.

In practice, the dependency refinement serves three main purposes.
From a semantic perspective, the added operators can be meaningful in the interpretation of the type system,
allowing distinct composition recipes for types which would otherwise be equated.
Further, through subsuming dependency label information, they allow for a backwards conversion into dependency-based syntactic frameworks.
Finally, from a parsing perspective, ILL$_\lolli$ by itself is word-order agnostic, meaning it admits more proofs than linguistically desired. 
The dependency decorations in this respect act as a balancing counter-weight, which on the one hand increases lexical type ambiguity,
but on the other hand provides valuable information to constrain proof search.

For the time being, we specify the dependency-decorated types at the lexical level of our provided proofs, but refrain from incorporating the structural rules imposed for reasons of brevity and simplicity.

\section{Extraction}
\label{section:extraction}
The formulation of the extraction process takes its inspiration from~\cite{DBLP:conf/lrec/MoortgatM02,moot2015type}, modulo some adaptations to account for the discrepancies in the type-logic and the source corpus.
Algorithmically, it may be perceived as a pipeline of three distinct components.
The first one concerns the transformation of the syntactically annotated input sentences into a \textit{directed acyclic graph} (DAG) format
that satisfies the input requirements of the remainder of the algorithm.
The intermediate one is responsible for assigning types to the DAG's nodes and asserting their validity within the phrase-local context.
Finally, the third component accepts the type-decorated DAG and identifies the interactions between its constituent types, thereby transforming it into a typelogical derivation.

\subsection{Lassy}
Lassy~\cite{vanNoord2013} is a dataset of syntactically annotated written Dutch. It is subdivided in two parts, Lassy Small and Lassy Large, both of which have been automatically annotated by the Alpino parser~\cite{bouma2001alpino}.
For the purposes of this work we focus our attention at the gold-standard Lassy Small, which (unlike its sibling) has been manually verified and corrected.
Lassy Small enumerates about 65\,000 sentences (or a million words), originating from various sources such as e-magazines, legal texts, manuals, Wikipedia content and newspaper articles among others.
The Lassy annotations are essentially DAGs, where nodes correspond to words and phrases labeled with lexical and phrasal categories, 
connected with edges that capture the syntactic functions between them\footnote{A concise description of the syntactic category
tags and dependency labels of Lassy can be found at \url{http://nederbooms.ccl.kuleuven.be/eng/tags} }. 
DAGs allow for reentrancy, whereby a node has multiple incoming dependency edges. 
In the Lassy annotation, cases of reentrancy are handled via phantom syntactic nodes coindexed with nodes that carry real content, so as to allow the annotation DAG to be visually represented as a tree.

\subsection{Preprocessing}
The extraction algorithm is formulated in terms of operations on DAGs. 
The first step mandated is therefore to collapse all coindexed duplicate nodes of each word or phrase into one, which inherits all the incoming edges of the original tree.

Next, we wish to treat annotation instances that are problematic for our type logic, stemming from schemes that under-specify the phrase structure.
The two usual culprits are discourse-level annotations, which do not exhibit a consistent function-argument articulation, and multi-word units, the children of which are syntactically indiscernible.
In the first case, we erase branches related to discourse structure (in many cases, these consist simply of structures annotated as ``discourse unit'' with substructures each marked as ``discourse part'').
To minimize the amount of annotations lost, we reconstruct an independent sample from each disconnected subgraph positioned under the cut-off point.
To resolve multi-word phrases without resorting to ad-hoc typing schemes, we merge participating nodes, essentially treating the entire phrase as a single word.
Punctuation symbols are dropped, as they are left untreated by the original annotation.

Beyond ensuring compatibility, we apply a number of transformations to Lassy's annotations designed to homo\-genize the types extracted. First, we remove the phantom syntactic nodes used to express the `understood' subject or object of non-finite verb forms, since this is semantic information that does not belong in a proper syntactic annotation. Second, we replace the generic \textit{conj} and \textit{mwu} tags (for conjunctions and multi-word units, respectively), by recursively performing a majority vote over their daughters' tags.
Finally, we refine the generic \textit{body} label by specifying the particular kind of clause it applies to, thus splitting it into \textit{rhd body}, \textit{whd body} and \textit{cmp body} for body of a relative, wh- and comparative clause respectively.

Edge erasures performed by the above procedures might lead to artifacts, which we take measures against.
We redeem DAGs with multiple sources by creating a distinct graph for each source, constituted by (a replica of) the source-reachable subset of the original graph. 
We remove redundant intermediate nodes with a single incoming and a single outgoing edge, and redirect the incoming edge to its corresponding descendant.
Conjunctions left with no more than one conjunct are truncated and replaced by their sole daughter.

\subsection{Type Assignment}
The output of the preprocessing procedure is a number of rooted DAGs per Lassy sample.
The goal now is to assign types to the nodes of these DAGs, according to the dependency-decorated scheme of Section~\secref{section:type_logic}.
We begin this endeavour by specifying two look-up tables that the type assignment algorithm is parametric to; one from part-of-speech tags and phrasal categories into atomic types, and one from dependency labels into modal operators (refer to Appendix \secref{subsec:trans}~for details).
Further, we specify the edge labels that correspond to phrasal heads and the ones that correspond to modifiers.
We then formulate the type assignment process as a conditional iteration over a DAG, with each iteration progressively inferring and assigning the types of a subset of the DAG's nodes, and the termination condition being the absence of any untyped nodes.

In abstract terms, the algorithm looks as follows:
\begin{algorithmic}
\State $\textsc{\textbf{TypeDag}}$ :: $\textsc{dag}$ $d$ $\to$ $\textsc{dag}$
\State \quad $\textsc{return}$ $\textsc{last}$ ($\textsc{unfold}$ $\textsc{Step}$ $d$)
\State $\textsc{\textbf{Step}}$ :: $\textsc{dag}$ $d$ $\to$ $\textsc{option}$ $\textsc{dag}$
\State \quad $d$' $\gets$ $\textsc{last}$ ($\textsc{unfold}$ $\textsc{TypeSimple}$ $d$)
\State \quad $d$' $\gets$ $\textsc{last}$ ($\textsc{unfold}$ $\textsc{TypeNonLocal}$ $d$')
\State \quad $d$' $\gets$ $\textsc{last}$ ($\textsc{unfold}$ $\textsc{TypeConjunctions}$ $d$')
\State \quad $\textsc{return}$ $d$' $\textsc{if}$ $d$' $\neq$ $d$ else $\textsc{Nothing}$
\end{algorithmic}

\begin{figure*}
    \centering
    \begin{subfigure}{\textwidth}
        \centering
        \includegraphics[scale=0.475]{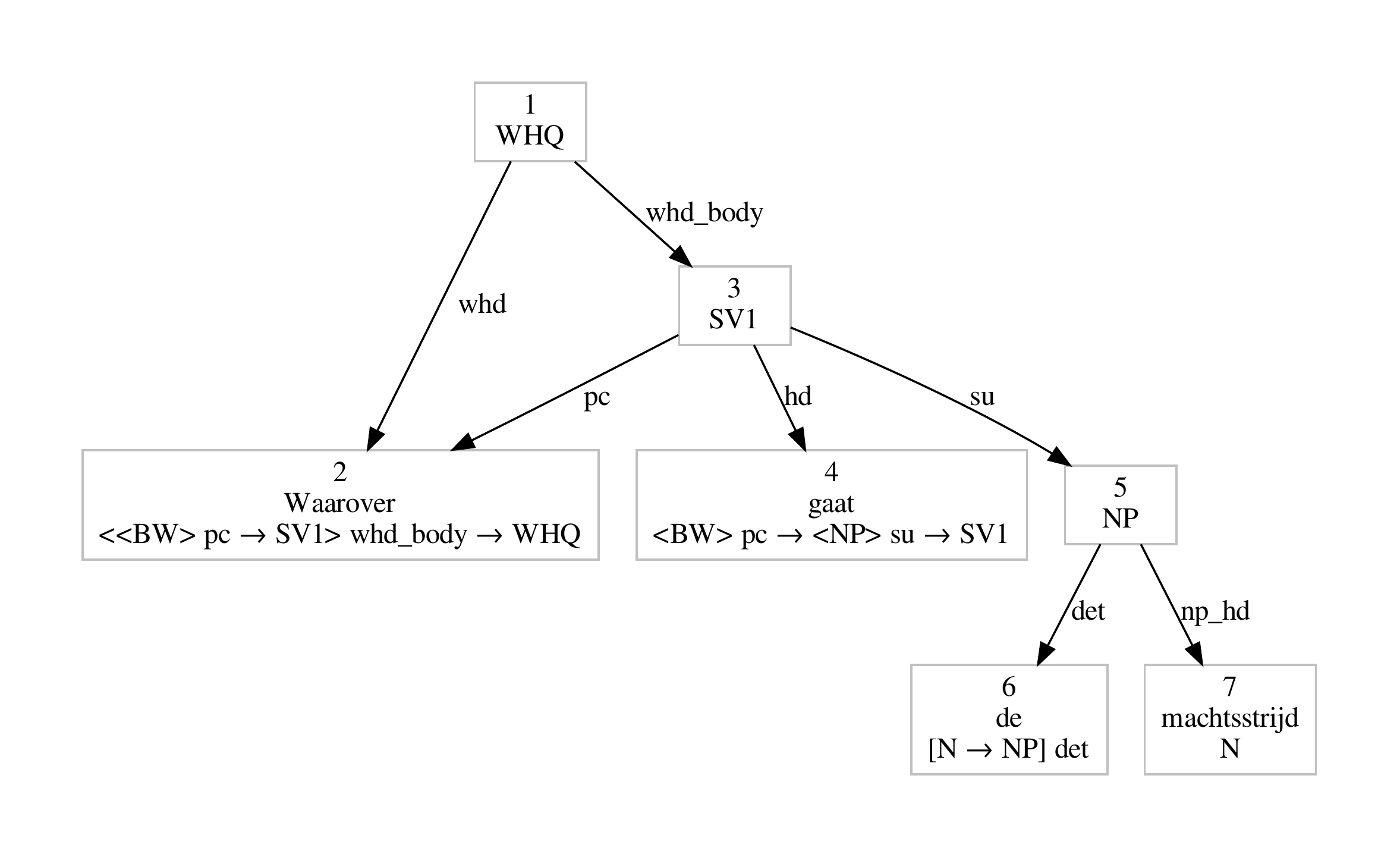}
        \caption{
        The dependency graph, as typed by the extraction component.
        The algorithm begins by simply translating the context-independent root 1 and leaf nodes 2 and 7 to $\type{WHQ}$, $\type{BW}$ and $\type{N}$ respectively.
        Given the absence of untyped children (modulo heads), intermediate nodes 3 and 5 are then typed as $\type{SV}1$ and $\type{NP}$, followed by the typing of functor nodes 4 and 6 as $\diamond^{\text{pc}}\type{BW}\li\diamond^{\text{su}}\type{NP}\li\type{SV}1$ and $\Box^{\text{det}}(\type{N}\li\type{NP})$.
        As 2 is also a phrasal head, its type is initially updated to $\diamond^{\text{whd body}}\type{SV}1 \li \type{WHQ}$.
        Finally, the second iteration step alters the type of 2 to account for the hypothetical gap, giving it its final assignment $\diamond^{\text{whd body}} (\diamond^{\text{pc}}\type{BW}\li \type{SV}1)\li\type{WHQ}$.
        }
        \label{subfig:example_dag}
    \end{subfigure}
    
    \begin{subfigure}{\textwidth}
        \centering
        \small
        \[
        \infer[\multimap E]{
        \text{Waarover} \left( \text{gaat} \ \left( \left( \text{de} \right)_\text{det} \ \text{machtstrijd} \right)_\text{su} \right)_{\text{whd body}} \vdash \type{WHQ}
        }{
        \infer[\mathcal{L}]{
            \vdash \diamond^{\text{whd body}} (\diamond^{\text{pc}}\type{BW}\li \type{SV}1)\li\type{WHQ}}{\text{Waarover}}
        &
        \hspace{-50pt}
        \infer[\diamond^{\text{whd body}} I]{
        \left( \text{g} \ \left( \left( \text{d} \right)_\text{det} \ \text{m} \right)_\text{su} \right)_{\text{whd body}} \vdash \diamond^{\text{wh body}}(\diamond^{\text{pc}}\type{BW}\li \type{SV}1)}{
        \infer[\multimap I]{
        \text{g} \ \left( \left( \text{d} \right)_\text{det} \ \text{m} \right)_\text{su} \vdash \diamond^{\text{pc}}\type{BW}\li \type{SV}1
        }{
        \infer[\multimap E]{
        \diamond^{\text{pc}}\type{BW}, \ \text{g} \ \left( \left( d \right)_\text{det} \ \text{m} \right)_\text{su} \vdash \type{SV}1}{
        \infer[\multimap E]{
        \diamond^{\text{pc}}\type{BW}, \ \text{gaat} \vdash \diamond^{\text{su}}\type{NP}\li\type{SV}1}{
        \infer[\mathcal{L}]{\vdash \diamond^{\text{pc}}\type{BW}\li\diamond^{\text{su}}\type{NP}\li\type{SV}1}{\text{gaat}}
        &
        \hspace{10pt}
        \infer[\mathcal{A}]{\diamond^{\text{pc}}\type{BW} \vdash \diamond^{\text{pc}}\type{BW}}{}
        }
        & 
        \infer[\diamond^{\text{su}} I]{
        \left( \left( \text{d} \right)_\text{det} \ \text{m} \right)_\text{su} \vdash \diamond^{\text{su}} \type{NP}}
        {\infer[\multimap E]{\left( \text{d} \right)_\text{det} \ \text{m} \vdash \type{NP}}{
        \infer[\Box^{\text{det}} I]{\left( \text{d} \right)_\text{det} \vdash \type{N} \li \type{NP}}
        {\infer[\mathcal{L}]{\vdash \Box^{\text{det}} \type{N} \li \type{NP}}{\text{de}}}
        &
        \infer[\mathcal{L}]{\vdash \type{N}}{\text{machtstrijd}}
        }
        }}}}}
        \]
        \caption{Structurally decorated linear logic corresponding to the above dependency graph, showcasing the effect of the $\diamond I$ and $\Box E$ rules on recovering the original dependency structure in the form of annotated bracketings. Intermediate steps depict words using their initial letters for space economy. Lexical type assignments are denoted with $\mathcal{L}$ and the identity axiom with $\mathcal{A}$.}
        \label{subfig:example_proof}
    \end{subfigure}
    \caption{Typed DAG and structurally annotated proof for Lassy sample \texttt{WS-U-E-A-0000000236.p.11.s.1.xml}, depicting an analysis for the phrase \textit{Waarover gaat de machtstrijd} (``What is the power struggle about?'').}
    \label{fig:example}
\end{figure*}

The iteration loop consists of three steps, each being the unfold of a function that takes a DAG, selects its fringe of typeable nodes, assigns a type to each node within, and returns the new (partially) typed DAG.
Each of these three functions differs in how it selects for its fringe and how it manufactures type assignments.
We detail their functionality in the next paragraphs; an illustrative example is also given in Figure~\ref{subfig:example_dag}.

\paragraph{Simple Clauses}
The first step internalizes the typing of sub-graphs within the DAG that exhibit simple syntactic clauses.

First off, we select \textit{context-independent} nodes, i.e. the root, plus those that are either leaves or have all of their daughters typed (possibly excluding modifier- and head-labeled daughters) and are not targets of exclusively modifying or head-labeled edges themselves; we type them by simply translating their part-of-speech or phrasal category tags into the corresponding atomic type.

The next step is to assign a type to nodes acting as \textit{phrasal heads} and \textit{modifiers}.
These may only be typed insofar as both their parents and all of their siblings are typed, imposing the analogous fringe conditions.
We assign phrasal heads the complex type:
\[
\diamond^{d_1} \type{A}_1 \li \dots \li \diamond^{d_n} \type{A}_n \li \type{R} \text{,}
\]
with $\diamond^{d_1}\type{A}_1, \dots , \diamond^{d_n} \type{A}_n$ the list of dependency-decorated types for the \textit{complements} and $\type{R}$ the phrasal type for the \textit{result}.
In order to obtain a consistent currying of multi-argument function types, we appeal to an obliqueness ordering of the dependency labels (see e.g.~\cite{dowty}),
and have the curried $n$-ary function type consume its $n$ arguments
from most oblique $\diamond^{d_1} \type{A}_1$ to least oblique $\diamond^{d_n} \type{A}_n$.
We refer to Appendix~\secref{subseq:obli}~for details of the obliqueness order assumed here.

\textit{Modifiers} are treated as non-head functors; their typing is based on the polymorphic scheme $\Box^{d} \left( \type{R} \li \type{R} \right)$, i.e. they are typed as endomorphisms of the phrasal type they are modifying, fixed to determine the dependency role $d$ they realize.

Similarly, \textit{determiners} are also treated as non-head functors: nouns are recognized as the syntactic heads of a noun phrase, in which determiners still assume the functor role. As such, they are assigned a boxed type $\Box^{\text{det}}(\type{N} \li \type{NP})$.

In order to obtain proper higher-order type assignments (with functors deriving functors), head (and determiner) nodes are typed in a bottom-up fashion, whereas modifier nodes are typed top-down.

\paragraph{Non Local Dependencies}
Lassy treats non-local phenomena such as relative clauses and constituent questions by inserting a \textit{secondary} edge pointing from a phrasal node embedded (possibly deeply) within the relative or question clause body on to the relativizing or interrogative pronoun.
Such pronouns then serve two roles.
In their primary role, they act as head of the relative or interrogative clause they project.
In their second role, they contribute to the composition of the dependency domain of the subclause where their secondary edge has its origin.
In this respect, they can be distinguished from other nodes due to having multiple incoming edges, the labels of which are distinct from one another.
Both of these two roles have already been addressed by the algorithm, but only in isolation.

To reconcile this, we select our fringe as nodes falling under this construction, and which have already been assigned some implicational type $\diamond^d\type{X} \li \type{Y}$ depicting the top-level clause functor (conforming to the flow of the first iteration step).
We then inspect the secondary dependency edge originating from (a subgraph of) $\type{X}$ in order
to update the aforementioned type to $\diamond^d(\diamond^e \type{E} \li \type{X}) \li \type{Y}$.

The updated higher-order type has a nested implication: from the parsing-as-deduction perspective this
means that in order to obtain a result of type $\type{Y}$, the relative or question pronoun has to assemble its argument $\type{X}$ with the aid of a hypothesis $\diamond^e \type{E}$.
Hypothetical reasoning (the $\lolli$ Introduction rule) is a key feature of our typelogical toolkit;
it obviates the need for phantom `gap' categories in unbounded dependency constructions. 

The identity of $\diamond^e$ and $\type{E}$ depends on what the label of the secondary dependency edge is. 
In case of an argument edge, it is simply the pair of translations from the dependency label and the node's part of speech or phrasal syntactic tag respectively.
In case of a modifying edge, the hypothesis subtype $\type{E}$ is itself a box functor $\Box^{m} \left( \type{Z} \li \type{Z} \right)$, corresponding to the endomorphism on the type $\type{Z}$ of the edge's source node.
In both cases, if the source node is in fact a sub-graph of $\type{X}$, the hypothesis type $\diamond^e \type{E}$ is prefixed with the structural $!$ operator, allowing it to traverse intermediate dependency domains as necessary to reach the point where ($\li I$) can withdraw it.

For any non-terminal nodes the type of which has been altered, we iteratively update the types of all heads and modifiers lying underneath in the DAG so as to account for the new phrasal type.

\paragraph{Conjunctions}
Phenomena of coordination and ellipsis pose a challenge for our resource-sensitive type logic: 
deriving conjunction types for incomplete phrases from the types of the complete conjuncts would require copying in the logic, an operation that our linear type system rules out.
Our approach here is to replace syntactic copying in the logical derivations by lexical polymorphism.
More precisely, \textit{coordinators} are typed according to the generically polymorphic scheme:
\[
\diamond^{\text{cnj}} \type{X} \li \dots \li \diamond^{\text{cnj}} \type{X} \li \type{X}\text{,}
\] where the number of arguments is concurring with the number of conjuncts.
To determine the value of $\type{X}$, we take the following steps. We begin by setting $\type{X}_0$ as the atomic type assigned to each of the conjuncts\footnote{We ensure a singular set of conjunct types by standardizing them using a majority-biased conjunct type relabeling.}.
We then select copied nodes (that is, nodes with multiple incoming edges, the labels of which coincide), which are lying under the inspected conjunction node and not below any other common ancestor and set $\type{X}$ to:
\[
! \diamond^{d_1} \type{C}_1 \li \dots \li ! \diamond^{d_n} \type{C}_n \li \type{X}_0 \text{,}
\]
where $\diamond^{d_1}\type{C}_1, \dots, \diamond^{d_n}\type{C}_n$ 
is the obliqueness-sorted list of dependency-decorated types of the copied nodes, combined with the structural $!$ to allow their unhindered relocation within the conjunct's dependency domain.
This scheme provides a uniform treatment of arbitrarily nested elliptical conjunctions and allows their logical derivation by means of higher-order types and hypothetical reasoning, without appealing to copying.


\subsection{Axiom Linking}
The algorithm specified above assigns a type to each DAG node; the multiset of types given to terminal nodes should admit the derivation of the root's type (i.e. the type of the sample phrase as a whole should be derivable by the types of its constituent words); it does not, however, specify the derivation itself.
To that end we design an additional algorithmic component, which accepts a type-annotated DAG and produces the linear logic proof it prescribes.

The first choice to make is of how to encode proofs; standard choices would be Gentzen style proofs (either in natural deduction or sequent format) or proofnets~\cite{girard1987linear}.
The proofnet presentation is more appealing, as it combines the pleasant property of natural deduction (one-to-one correspondence
with the program terms for meaning composition) with the good computational properties of sequent calculus (decidable proof search).
The type system's structural rules are not explicitly specified in the proof representations.

Providing the full theory behind proofnets goes beyond the scope of this paper; what follows is a simplified summary.
We begin by assigning types a \textit{polarity}. 
In the context of a logical judgement, types appearing in antecedent position (i.e. assumptions left of the turnstile) are negative, whereas types appearing in succedent position (i.e. conclusions right of the turnstile) are positive. 
Polarities are then propagated to subformulas as follows.
If a type is atomic, its polarity remains unaltered. 
If it is an implication $\type{T}_1 \li \type{T}_2$, then the polariy of $\type{T}_2$ persists, whereas the polarity of $\type{T}_1$ is reversed.

Atoms nested within complex types are assigned a polarity by recursive application of the above scheme.
At its essence, a proofnet is a bijection between positive and negative atoms, i.e. a pairing of each positive atom with an (otherwise equal) negative one\footnote{This bijection must satisfy certain correctness criteria, i.e. not all such bijections constitute valid proofnets.}.

Finding a proof is then equated with constructing the appropriate bijection.
To ensure that such a bijection is indeed possible, we first perform a rudimentary correctness check, asserting the branch-wise invariance count of atoms and implications~\cite{van1991language}.
We then project the types of the DAG's terminal nodes into a flattened sequence, sort them based on the corresponding word order, and decorate each atom with an integer, thus distinguishing between unique occurrences of the same atom.
Our goal then lies in propagating these indices upwards along the DAG, linking atom pairs as we go.
The algorithmic procedure is outlined below.

We first instantiate an empty proofnet in the form of a bijective function from indices to indices. 
We additionally implement a function, which takes a proofnet and two equal types of inverse polarity and recursively matches their corresponding atoms, updating the proofnet in the process.
For simple branchings, we isolate the arguments of the phrasal functor and identify them with the types of its sibling nodes on the basis of the diamond or box decoration of the former and the dependency labels of the latter.
The resulting pairs are matched and the proofnet is expanded.
The type of the node dominating the branch is then indexed with the functor's result index(es).
A bottom-up iteration then gradually indexes the DAG's non-terminal nodes while filling in the proofnet.

\begin{figure*}[t]
\floatbox[{\capbeside\thisfloatsetup{capbesideposition={right, top},capbesidewidth=0.3\linewidth}}]{figure}[\FBwidth]
{\caption{Fraction of words covered as a function of types included.
Y axis depicts the percentage of words that can be analyzed (\textit{logit}-transformed). X axis illustrates the percentage of unique types considered. A point $(x, y)$ then represents the $\%$ of words $y$ in the corpus that could be type-assigned if all but the $x\%$ most common types were discarded.}\label{fig:type_coverage}}
{\includegraphics[trim=2cm 2cm 4cm 2cm, clip, width=9cm, height=4.5cm]{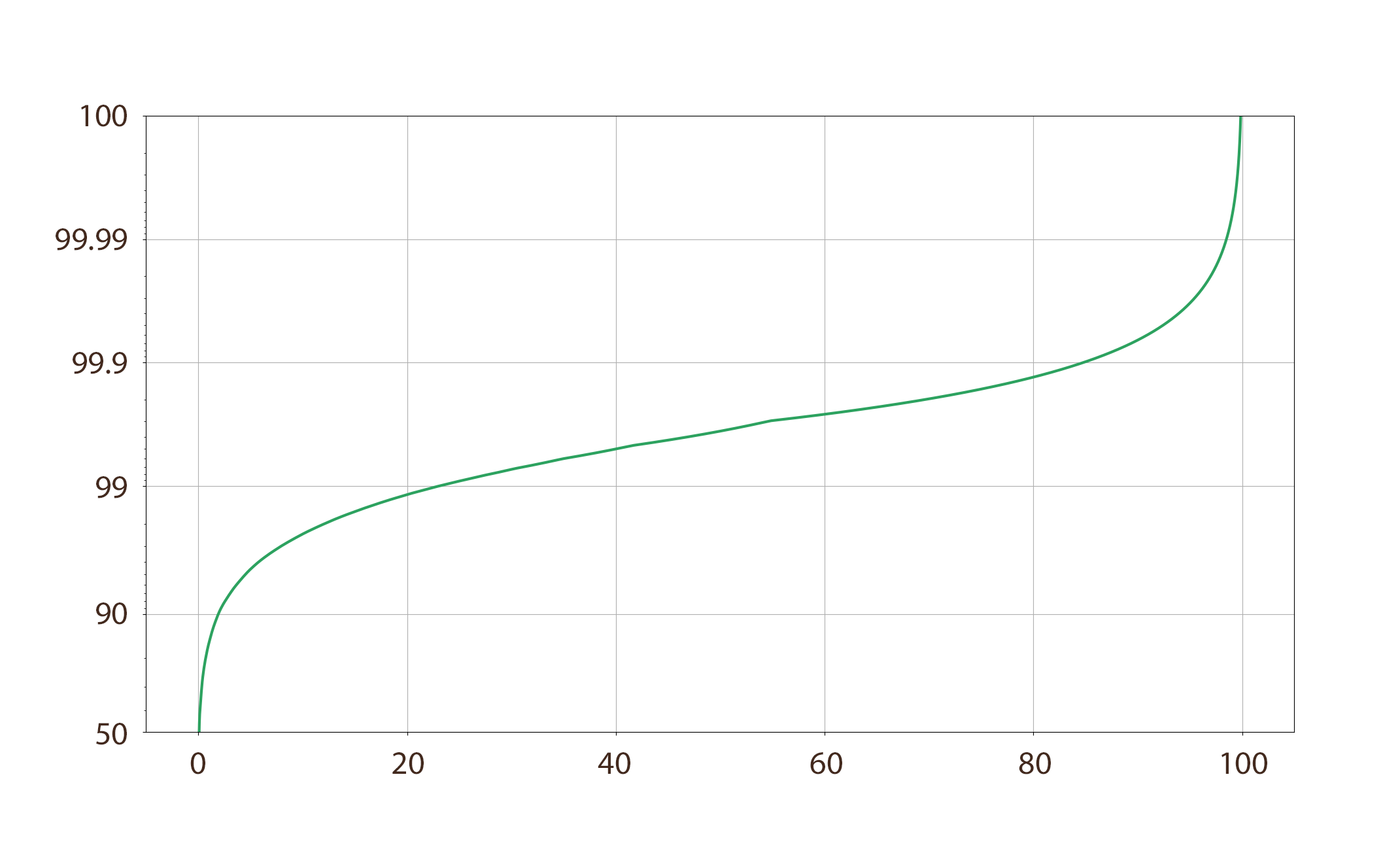}}
\end{figure*}

\begin{figure*}[t]
\floatbox[{\capbeside\thisfloatsetup{capbesideposition={right, top},capbesidewidth=0.3\linewidth}}]{figure}[\FBwidth]
{\caption{Lexical ambiguity histogram.
Y axis depicts the number of words (\textit{log}-scaled), X axis depicts the number of types (\textit{log}\textsubscript{2}-scaled).
A bar spanning the horizontal range $(x_1, x_2)$ with height $y$ indicates that a total of $y$ words are associated with $x_1$ to $x_2$ number of types.}\label{fig:type_ambiguity}}
{\includegraphics[trim=2cm 2cm 4cm 2cm, clip, width=9cm, height=4.5cm]{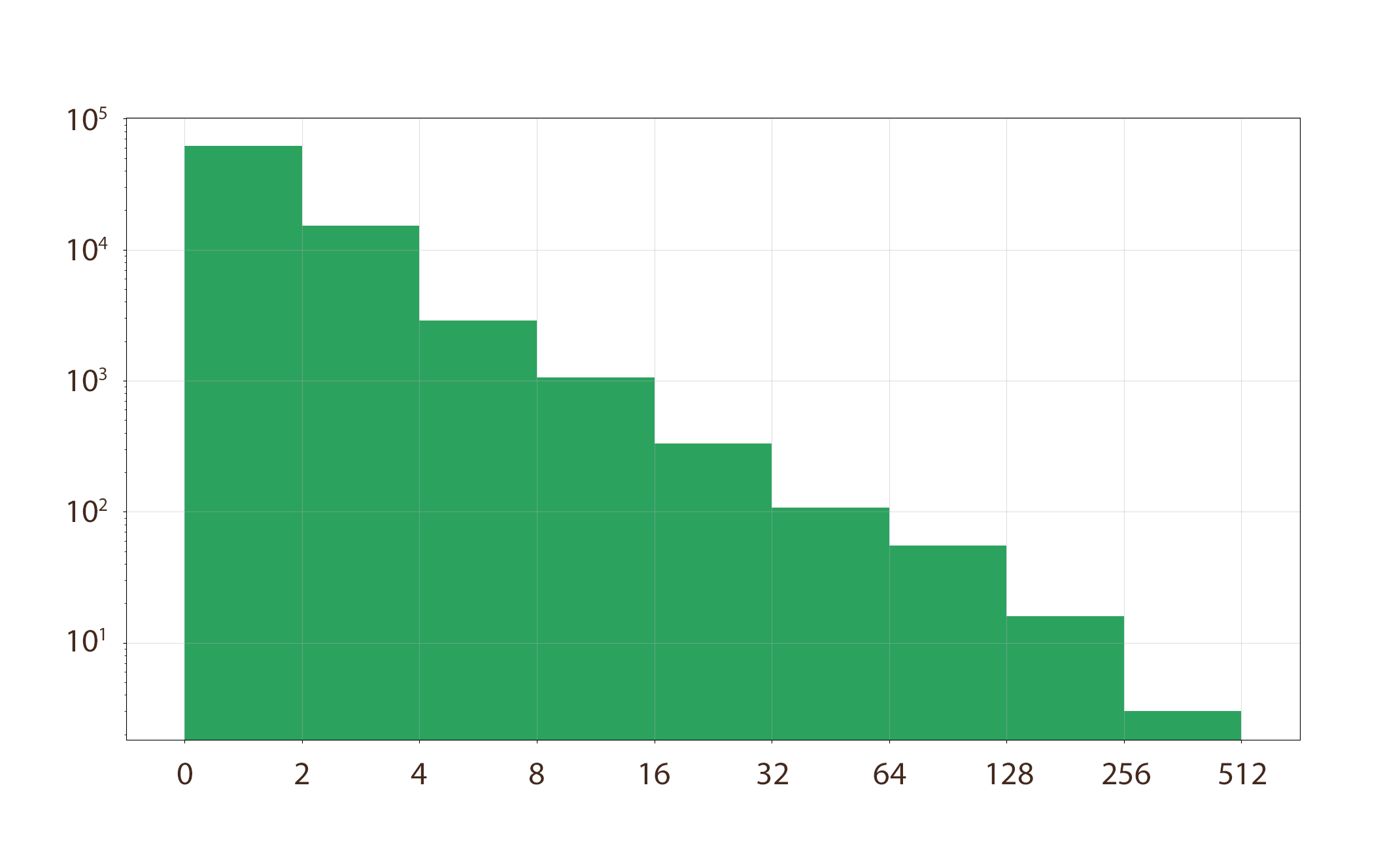}}
\end{figure*}

For constructions involving hypothetical reasoning, such as elliptic conjunctions and non-local phenomena, the process is a bit more intricate.
The higher-order types involved in these do double duty, both providing the functor that composes the outer phrase and supplying the material missing from the inner phrase(s).
To resolve such constructions, we simplify the higher-order types by subtracting their embedded arguments and traversing the DAG to find the branch that misses them.
Once there, we detach edges that are secondary or point to copied nodes, and replace them by edges (of the same label) that point to placeholder nodes carrying the aforementioned subtracted arguments.
This transformation essentially converts the typed DAG back into a typed tree, reducing the problem once more to the simple case.

\subsection{Verification}
The extraction procedure described in the previous section produces type assignments together with axiom links, i.e.~a bijection matching each atomic subformula occurrence to another of opposite polarity.
Albeit being legitimate bijections, these axiom links are not necessarily proof nets; a set of correctness criteria needs to be satisfied for a bijection to be translatable into a proof net (a proof net then typically corresponds to many proofs, which differ only in inessential rule permutations).
To validate the correctness of the algorithm, we use use LinearOne\footnote{Available at \url{https://github.com/RichardMoot/LinearOne}.}, a linear logic theorem prover to verify that all extracted structures are indeed linear logic proofs. 
In addition to asserting correctness and providing a sanity check on the extraction algorithm, LinearOne also produces a linear logic proof in sequent calculus and natural deduction presentations as well as the corresponding $\lambda$-term for each input proof net candidate.

\subsection{Analysis}
The end-to-end pipeline yields a number of samples, each corresponding to a mechanically verified typelogical derivation for a sentential or phrasal syntactic analysis.
Even though everything on the output end is proven correct (i.e. all extracted proof nets satisfy the correctness criteria), the extraction algorithm fails to produce a derivation in a limited number of instances.
These failures are recognized during runtime and their cause is pinpointed; we provide a breakdown of the algorithm's coverage at each step of the process.

\begin{figure*}[t]
\floatbox[{\capbeside\thisfloatsetup{capbesideposition={left, top},capbesidewidth=0.3\linewidth}}]{figure}[\FBwidth]
{\caption{Sentence length histogram.
Y axis depicts the number of sentences (\textit{log}-transformed), X axis depicts the number of words.
A bar spanning the horizontal range $(x_1, x_2)$ with height $y$ indicates that a total of $y$ sentences have a length of $x_1$ to $x_2$ words (with punctuations removed and multi-word phrases counted as a single lexical item).}\label{fig:lens}}
{\includegraphics[trim=2cm 2cm 4cm 2cm, clip, width=9cm, height=4.5cm]{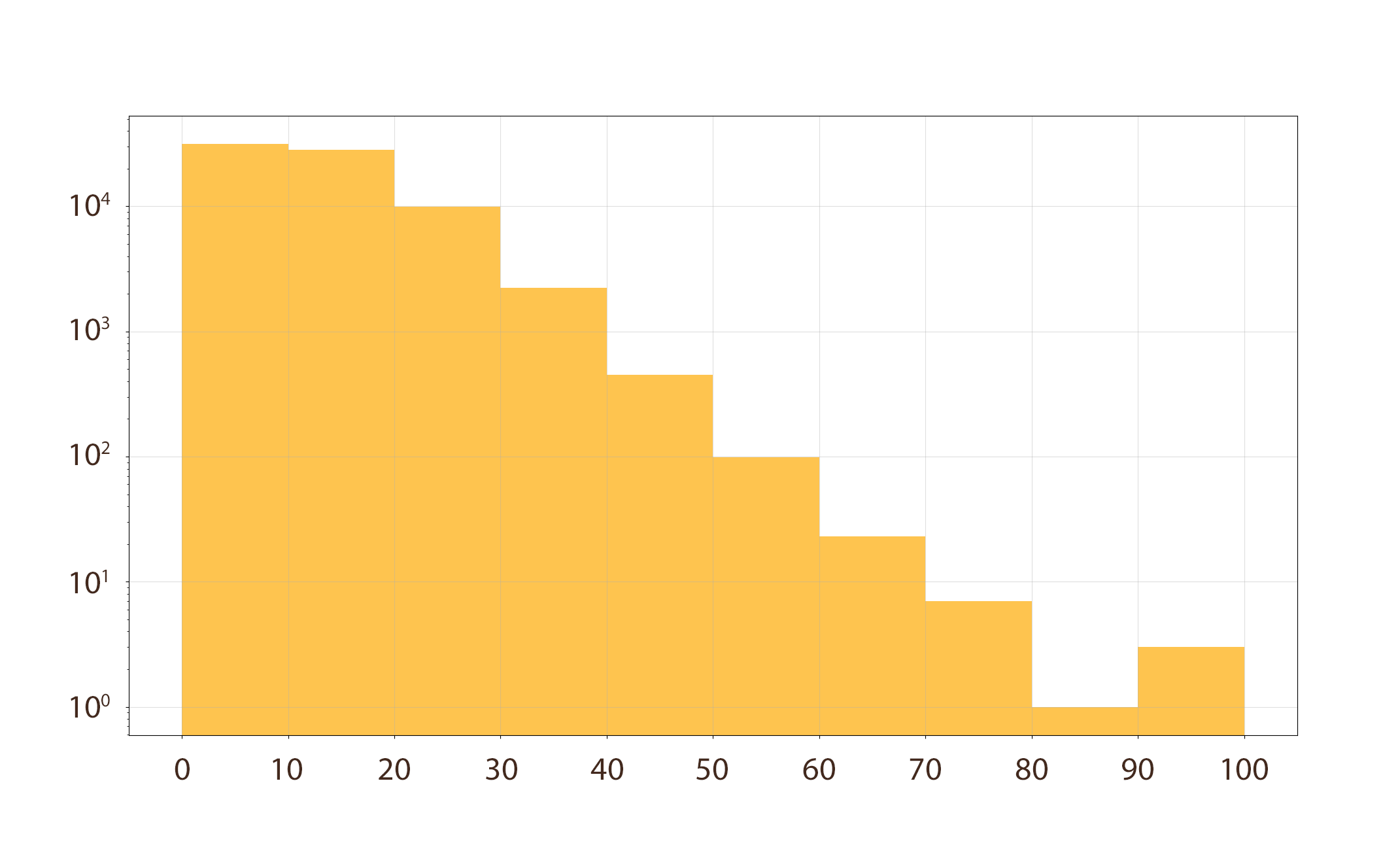}}
\end{figure*}

\begin{figure*}[t]
\floatbox[{\capbeside\thisfloatsetup{capbesideposition={left, top},capbesidewidth=0.3\linewidth}}]{figure}[\FBwidth]
{\caption{Fraction of sentences covered as a function of types included.
Y axis depicts the percentage of sentences that can be analyzed (\textit{logit}-transformed). X axis depicts the percentage of unique types considered. A point $(x,y)$ then represents the $\%$ of sentences $y$ in the corpus, all words of which could be type assigned if all but the $x\%$ most common types were discarded.}\label{fig:sentence_coverage}}
{\includegraphics[trim=0.5cm 2cm 4cm 2cm, clip, width=9.25cm, height=4.5cm]{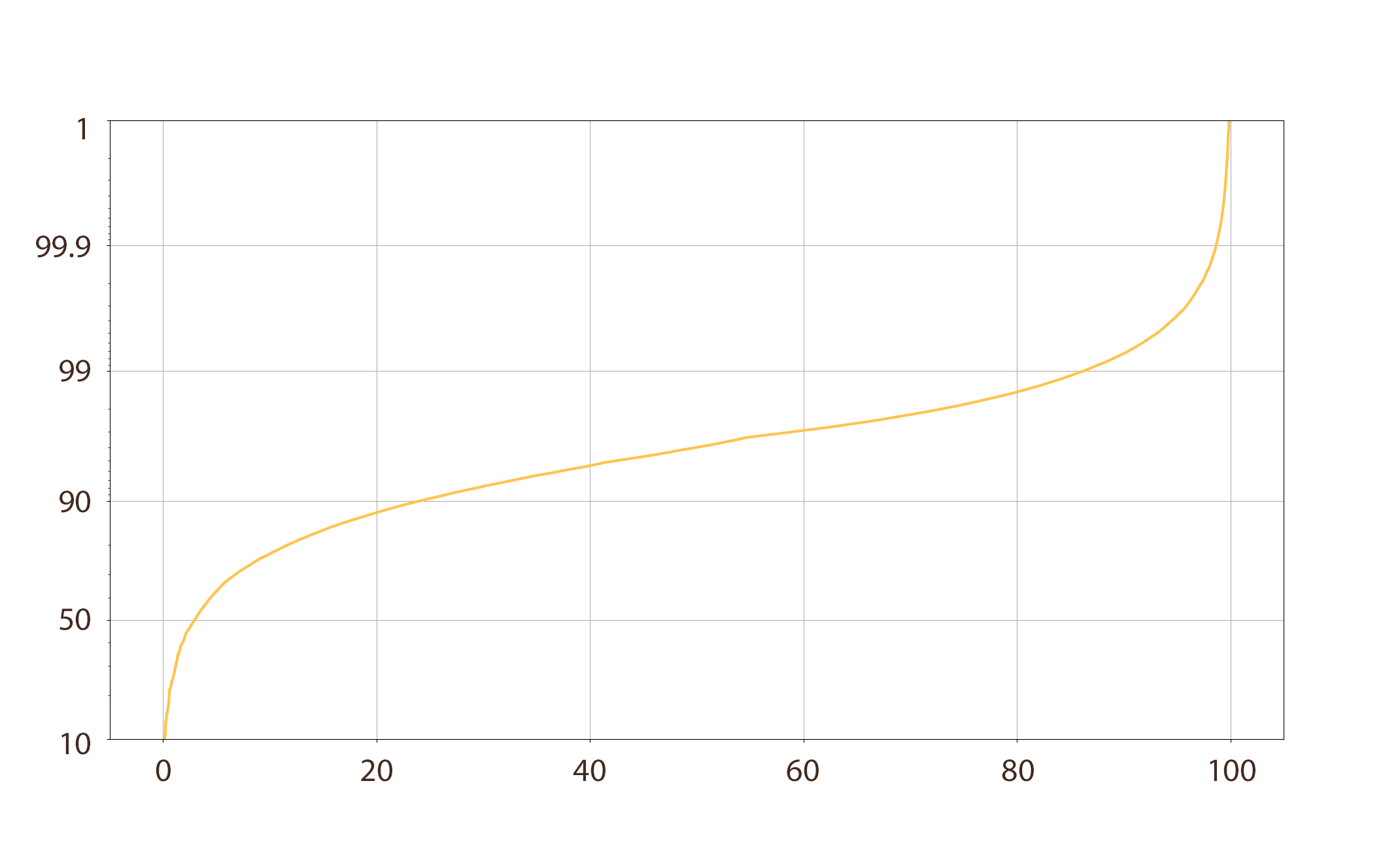}}
\end{figure*}

\paragraph{Preprocessing}
Lassy-Small contains 65\,200 analyses.
We drop 156 of these due to being a single punctuation mark.
Out of the remaining 65\,044, we obtain 75\,060 independent DAGs (1.15 DAGs per Lassy sample on average).

\paragraph{Typing}
The type assignment algorithm produces a correct output for 72\,198 samples (96.2\% input coverage).
The majority of failed cases relates to conjunction schemata; in particular, 1\,492 samples contain conjunction branches that lack a coordinating word and 443 samples contain asymmetric conjunctions.
Both cases could be trivially solved, for instance by promoting the first conjunct to the branch head or expanding the polymorphic coordinator scheme to account for unequal constituents; the ad-hoc nature of such solutions would however degrade the quality and consistency of the lexical types, and we therefore abstain from implementing them.
The remaining 494 cases (0.6\% of the input) are generic failures arising from copying outside the context of a conjunction, annotation errors and discrepancies, and preprocessing artifacts. 

\paragraph{Linking and Verifying}
Out of the 72\,198 typed DAGs, the axiom linking algorithm fails to produce a sane bijection in 6 (of which 4 are edge cases and 1 is an annotation error).
All 72\,192 outputs of the axiom linking algorithm are validated by the theorem prover, leading to an end-to-end coverage of 96.2\%.

\section{Resources}
\label{section:resources}
\subsection{Code}
We make the Python code implementing the extraction algorithm publicly available\footnote{The code can be found at \url{https://github.com/konstantinosKokos/Lassy-TLG-Extraction}.}.
It is parametric and tunable with respect to the part of speech and phrasal category translation tables (both in terms of domain and codomain, allowing either a refinement or a consolidation of the atomic type set considered), but also the dependency labels (similarly allowing an adjustment of the number of diamond operators).
The algorithm is, to a certain degree, agnostic about the underlying type system specification; in other words, it is easy to adapt to different grammars, and can be fine-tuned according to the user's needs and purposes.
Further, the algorithm is immediately applicable to Lassy-Large, which boasts a total of 700 million words, significantly outnumbering Lassy-Small.
Its massive size, together with the extraction algorithm, grants easy access to a large amount of cross-domain silver standard type-theoretic analyses, as well as a potential corpus for unsupervised language modeling enhanced with lexicalized structural biases.
Last but not least, the extraction algorithm is compatible with the Alpino parser's~\cite{bouma2001alpino} output format; combining the two would then essentially account to an ``off the shelf'' typelogical theorem prover for written Dutch that produces computational terms along with the standard parses.

\subsection{Lexicon}
Dissociating leaf nodes from their DAG, we obtain a weighted lexicon mapping words to type occurrences.
Our lexicon enumerates a total of 77\,283 distinct words and 5\,747 unique types made out of 31 atomic types and 26 dependency decorations.
On average, each word is associated with 1.79 types and each type with 24 words.
Figure~\ref{fig:type_ambiguity}~ presents a histogram of lexical type ambiguity; most words are unambiguous, being always assigned a single type, whereas 20 words (mostly coordinators and auxiliary verbs) are highly ambiguous, being associated with more than 128 unique types each.
Figure~\ref{fig:type_coverage}~ presents the relation between lexical coverage and types considered; evidently, the 1\,150 most common types (20\%) suffice to cover 99\% of the corpus (on a per-word basis).
The lexicon can be utilized as a stand-alone resource, useful for studying grammatical relations and syntactic variation at the lexical level.

\subsection{Theorems}
The core resource is a collection of 72\,192 typelogical derivations\footnote{The public subset of the dataset consists of 7\,924 sentences and is available online \url{https://github.com/konstantinosKokos/aethel-public} }.

The primitive component behind each derivation is a type-annotated sentence or phrase.
On average, samples consist of 12 lexical items (a 25\% drop compared to the unprocessed source corpus due to multi-word merges and detached branches); figure~\ref{fig:lens} presents a histogram of sample lengths.
As already hinted by Figure~\ref{fig:type_coverage}, the fine-grained nature of the type system has the side-effect of enlarging the lexicon's co-domain, and therefore the type sparsity, in comparison to other lexicalized grammar formalisms.
Figure~\ref{fig:sentence_coverage} presents the relation between corpus coverage and types considered; no less than the 5\,000 most common types (85\%) are required to parse 99\% of the corpus.
This suggests that, in themselves, the annotated sentences constitute a challenging supertagging task as well as a potential benchmark for open-world classification~\cite{cts}.

Of higher importance are, however, the derivations themselves.
As stated earlier, they are provided in four distinct formats; as natural deduction and sequent calculus proofs, bijections between atomic formulas (proofnets) as well as $\lambda$-terms.
The type system, being based on ILL, is agnostic about word order and thus inherently ambiguous. 
Out of the (possibly many) potential proofs, the dataset specifies the one that is linguistically acceptable, determining the correct flow of information within the sentence's constituents (an example of such a derivation is presented in Figure~\ref{subfig:example_proof}).
This has multiple ramifications and usecases. 

First and foremost, it opens a path towards integrated neuro-symbolic approaches to parsing. 
On one hand, lexical interactions are constrained to just those that are respectful of the typing information.
On the other hand, neural approaches can be applied to narrow down the resulting search space, simultaneously utilizing semantic and syntactic information, with the dependency enrichment providing an additional heuristic.
In this context, selecting a parsing methodology and the appropriate proof format is up to the end-user; for instance, shift-reduce parsing would be easier accomplished on the binary branching natural deduction structures~\cite{shieber1995principles,ambati2016shift}, $\lambda$-terms would be more accommodating for generalized translation architectures like sequence transducers~\cite{zettlemoyer2012learning}, proofnet bijections could be obtained via neural permutation learning ~\cite{mena2018learning} etc; supertagging and type representations can also be jointly optimized~\cite{lewis2016lstm,kasai2017tag}.

The dataset can also emerge as a useful resource for ``pure'' parsing as deduction. 
The type grammar and its derivations can be utilized as a stepping stone towards stricter type systems.
Noting that our abstract types are in alignment with Lambek types (modulo directionality), linear implications can be gradually replaced with directed divisions based on their aggregated corpus-wide behavior, easing the transition towards an (either hybrid or multi-modal) typelogical dataset.

\section{Conclusion}
\label{section:conclusion}
We have described a linear type system that captures abstract syntactic function-argument relations, but is also able to distinguish between arguments on the basis of their dependency roles.
We have presented a methodology for extracting these linear types as well as their interactions out of dependency parsed treebanks.
Our approach is modular, allowing a large degree of parameterization, and general enough to accommodate alternative type systems and source corpora.
We implemented and applied a concrete algorithmic instantiation, which we ran on the Lassy treebank, generating a large dataset of type-theoretic syntactic derivations for written Dutch.
Utilizing a theorem prover, we verified the correctness of the algorithm's output, and transformed it into a number of different realizations to facilitate its use in different contexts.
Taking advantage of the Curry-Howard correspondence between linear logic and the simply-typed $\lambda$-calculus, we naturally equated our derivations to computational terms which characterize the flow of information within a sentence.
Disconnecting types from their derivations, we are left with a pairing of sentences to type sequences; disconnecting them from their surrounding context, we obtain a highly refined lexicon mapping words to type occurrences.
We make a significant portion of the above resources publicly available.
Our hope is that our resources will find meaningful applications at the intersection of formal and data-driven methods, in turn giving rise to practically applicable insights on the syntax-semantics interface.

\section*{Acknowledgements}
Konstantinos and Michael are supported by the Dutch Research Council (NWO) under the scope of the project ``A composition calculus for vector-based semantic modelling with a localization for Dutch'' (360-89-070).

\section*{Bibliographical References}
\label{main:ref}

\bibliographystyle{lrec}
\bibliography{main}


\clearpage

\appendix

\section*{Appendix}
\label{sec:appendix}

\subsection{Translation Tables}
\label{subsec:trans}
\paragraph{Atomic Types}
Table~\ref{table:lex}~presents the set of atomic types and their origins (part-of-speech and phrasal category tags).
The current translation domain utilizes the Lassy part of speech tagset (\textit{pt}); other options could be either the alpino tagset (\textit{pos}) or even the detailed tagset (\textit{postag}).
Note also that in our usecase there is a one-to-one correspondence between tags and types.
This does not necessarily need to be the case; one could as well choose to collapse one or more tags onto the same type (e.g. translate \textit{vnw} to $\type{np}$ or all sentential tags to $\type{s}$).
The extraction algorithm is parametric to all above possible variations.

\begin{table}[ht]
\begin{center}
\begin{tabularx}{1\linewidth}{uss}
      \textbf{Tag} & \textbf{Description} & \textbf{Assigned Type}\\
      \toprule
      \multicolumn{3}{c}{Short POS Tags}\\
      \midrule[0.005pt]
            
      \textit{adj} & Adjective & \type{adj}\\
      \textit{bw} & Adverb & \type{bw}\\
      \textit{let} & Punctuation & \type{let}\\
      \textit{lid} & Article & \type{lid}\\
      \textit{n} & Noun & \type{n}\\
      \textit{spec} & Special Token & \type{spec}\\
      \textit{tsw} & Interjection & \type{tsw}\\
      \textit{tw} & Numeral & \type{tw}\\
      \textit{vg} & Conjunction & \type{vg}\\
      \textit{vnw} & Pronoun & \type{vnw}\\
      \textit{vz} & Preposition & \type{vz}\\
      \textit{ww} & Verb & \type{ww}\\
      \midrule[0.005pt]
      \multicolumn{3}{c}{Phrasal Category Tags}\\
      \midrule[0.005pt]
      \textit{advp} & Adverbial Phrase & \type{adv}\\
      \textit{ahi} & Aan-Het Infinitive & \type{ahi}\\
      \textit{ap} & Adjectival Phrase & \type{ap}\\
      \textit{cp} & Complementizer Phrase & \type{cp}\\
      \textit{detp} & Determiner Phrase & \type{detp}\\
      \textit{inf} & Bare Infinitival Phrase & \type{inf}\\
      \textit{np} & Noun Phrase & \type{np}\\
      \textit{oti} & Om-Te Infinitive & \type{oti}\\
      \textit{pp} & Prepositional Phrase & \type{pp}\\
      \textit{ppart} & Past Participial Phrase & \type{ppart}\\
      \text{ppres} & Present Participial Phrase & \type{ppres}\\
      \textit{rel} & Relative Clause & \type{rel}\\
      \textit{smain} & SVO Clause &\type{smain}\\
      \textit{ssub} & SOV Clause & \type{ssub}\\
      \textit{sv1} & VSO Clause & \type{sv1}\\
      \textit{svan} & Van Clause & \type{svan}\\
      \textit{ti} & Te Infinitive & \type{ti}\\
      \textit{whq} & Main WH-Q & \type{whq}\\
      \textit{whrel} & Free Relative & \type{whrel}\\
      \textit{whsub} & Subordinate WH-Q & \type{whsub}\\
      \midrule[0.005pt]
\end{tabularx}
\end{center}
\caption{Part-of-speech tags and phrasal categories, and their corresponding type translations.}
\label{table:lex}
\end{table}

\paragraph{Dependency Decorations}
Table~\ref{table:colors}~ presents the set of dependency decorations (i.e. modal operators) and their descriptions.
Most decorations coincide with a Lassy dependency label or a derivative thereof.

\begin{table}
\centering 
\begin{tabularx}{1\linewidth}{scu}
      \textbf{Decoration} & \textbf{Description} & \textbf{Precedence}\\
      \toprule
      \textit{app} & Apposition & 20\\
      \textit{cmp\_body} & Complementizer body & 18\\
      \textit{cnj} & Conjunct & 0\\
      \textit{det} & Noun-phrase determiner & 1\\
      \textit{hdf} & Final part of circumposition & 7\\
      \textit{ld} & Locative Complement & 6\\
      \textit{me} & Measure Complement & 5\\
      \textit{mod} & Modifier & 21\\
      \textit{obcomp} & Comparison Complement & 3\\
      \textit{obj1} & Direct Object & 12\\
      \textit{obj2} & Secondary Object & 10\\
      \textit{pc} & Prepositional Complement & 8\\      
      \textit{pobj} & Preliminary Object & 13\\      
      \textit{predc} & Predicative Complement & 11\\      
      \textit{predm} & Predicative Modifier & 19\\
      \textit{rhd-body} & Relative clause body & 17\\
      \textit{se} & Obligatory Reflexive Object & 9\\      
      \textit{su} & Subject & 14\\
      \textit{sup} & Preliminary Subject & 15\\
      \textit{svp} & Separable Verbal Participle & 2\\   
      \textit{vc} & Verbal Complement & 4\\
      \textit{whd-body} & WH-question body & 16\\
\end{tabularx}
\caption{Dependency relations and their corresponding implication labels.}
\label{table:colors}
\end{table}

\subsection{Obliqueness Hierarchy}
\label{subseq:obli}
Phrasal heads are assigned functor types; in the multi-argument case, these would be of the form:
\[
(\type{A}_1 \otimes \type{A}_2 \otimes \dots \type{A}_n) \li \type{R}
\text{,}\]
or their curried equivalent:
\[
\type{A}_1 \li \type{A}_2 \li \dots \type{A}_n \li \type{R}
\]

To avoid the inconvenience of (in this case, superficial) distinction between different argument type permutations, we impose a strict full order on argument sequences, loosely based on the obliqueness hierarchy of their syntactic roles (apparent through their modal decorations).
The ordering is presented in the third column of Table~\ref{table:colors}; the lower a label's number, the less prominent (i.e.~more oblique) the argument it marks, causing it to be consumed first.
This scheme, recursively applied, provides a unique implicational type for each functor's argument-permutation class.
Functors carrying \textit{cnj}-decorated arguments (i.e. coordinators and their derivatives) are the only kind of functor which permits two distinct argument of the same decoration; we sort those based on the conjuncts' order within the sentence.

\end{document}